\documentclass[10pt,twocolumn,letterpaper]{article}

\usepackage[pagenumbers]{iccv} 

\usepackage[accsupp]{axessibility}  

\definecolor{iccvblue}{rgb}{0.21,0.49,0.74}
\usepackage[pagebackref,breaklinks,colorlinks,allcolors=iccvblue]{hyperref}
\usepackage{amsmath}
\usepackage{bm}

\def\paperID{11278} 
\def\confName{ICCV}
\def\confYear{2025}

\title{\textsc{ShortFT}: Diffusion Model Alignment via Shortcut-based Fine-Tuning}

\author{
{\bf Xiefan Guo\textsuperscript{1,2}\quad 
Miaomiao Cui\quad 
Liefeng Bo\quad 
Di Huang\textsuperscript{1,2}\footnotemark[1]}\\[5pt]
{\normalsize \textsuperscript{1}State Key Laboratory of Complex and Critical Software Environment, Beihang University, Beijing 100191, China}\\
{\normalsize \textsuperscript{2}School of Computer Science and Engineering, Beihang University, Beijing 100191, China}\\
{\tt\small \{xfguo,dhuang\}@buaa.edu.cn}\\
{\normalsize Project: \url{https://xiefan-guo.github.io/shortft}}
}

\begin{document}

\twocolumn[{
\renewcommand\twocolumn[1][]{#1}
\maketitle
\vspace{-0.5cm}
\begin{center}
\centering
\setlength{\abovecaptionskip}{0.1cm}
\captionsetup{type=figure}
\includegraphics[width=1.\linewidth]{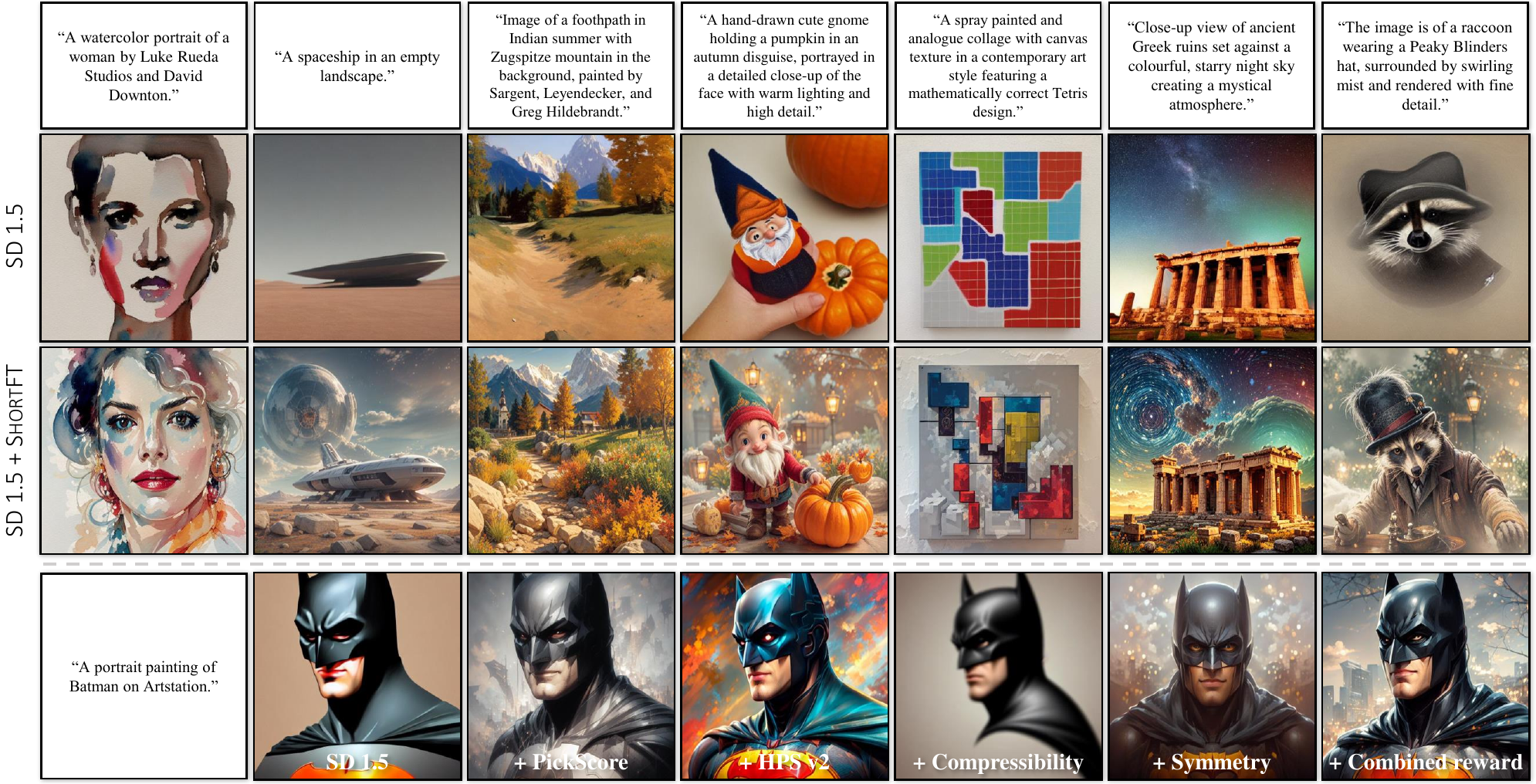}
\captionof{figure}{\textbf{Example results synthesized by \textsc{ShortFT}}. \textsc{ShortFT} endeavors to achieve the alignment of diffusion models with reward functions by facilitating the end-to-end backpropagation of the targeted reward gradient throughout the denoising chain. Our method has exhibited remarkable efficacy, particularly evident in the realms of text-image alignment and the overall enhancement of image quality (Top). Moreover, its versatility is underscored by the successful application across diverse reward functions, substantially amplifying alignment performance (Bottom). Combined reward is a weighted combination of rewards: PickScore = 10, HPS v2 = 2, Aesthetic = 0.05.}
\label{fig:teaser}
\end{center}
}]

\footnotetext[1]{Corresponding author.}

\begin{abstract}
Backpropagation-based approaches aim to align diffusion models with reward functions through end-to-end backpropagation of the reward gradient within the denoising chain, offering a promising perspective. However, due to the computational costs and the risk of gradient explosion associated with the lengthy denoising chain, existing approaches struggle to achieve complete gradient backpropagation, leading to suboptimal results. In this paper, we introduce Shortcut-based Fine-Tuning (\textsc{ShortFT}), an efficient fine-tuning strategy that utilizes the shorter denoising chain. More specifically, we employ the recently researched trajectory-preserving few-step diffusion model, which enables a shortcut over the original denoising chain, and construct a shortcut-based denoising chain of shorter length. The optimization on this chain notably enhances the efficiency and effectiveness of fine-tuning the foundational model. Our method has been rigorously tested and can be effectively applied to various reward functions, significantly improving alignment performance and surpassing state-of-the-art alternatives.
\end{abstract}

\begin{figure*}
\centering
\setlength{\belowcaptionskip}{-0.1cm}
\setlength{\abovecaptionskip}{0.2cm}
\includegraphics[width=0.98\linewidth]{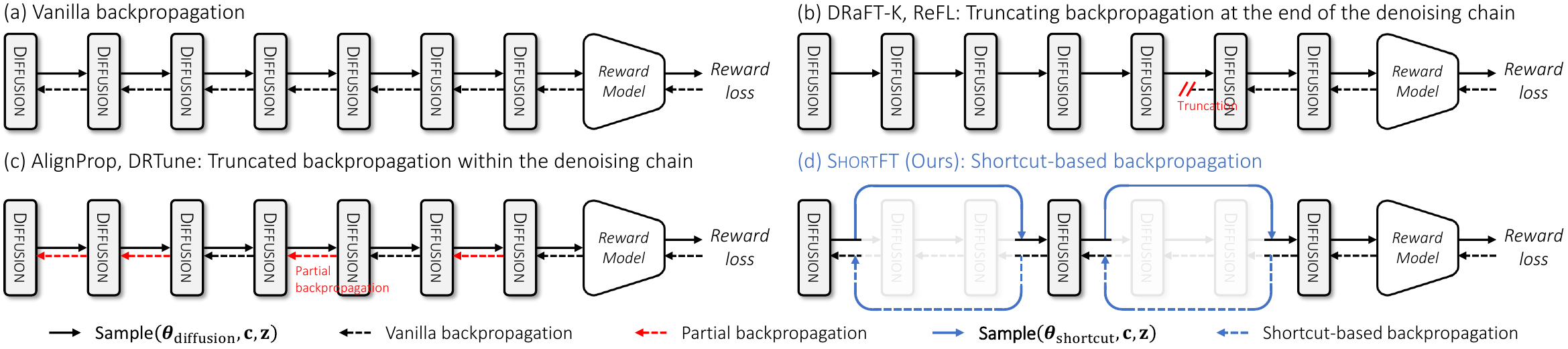}
\caption{\textbf{Comparison of fine-tuning strategies.} (a) The vanilla backpropagation-based fine-tuning strategy, which suffers from lengthy backpropagation chains. (b) DRaFT-K \cite{clark2024directly} and ReFL \cite{xu2024imagereward} truncate the backpropagation chain, focusing on the latter half of the denoising chain, they ignore the direct supervision at the early stage, resulting in suboptimal alignment with text prompts. (c) AlignProp \cite{prabhudesai2023aligning} and DRTune \cite{wu2024deep} truncate part of the backpropagation within the denoising chain by disabling some gradients. Specifically, given the denoising operation: $\mathbf{x}_{t-1} = \textcolor{red}{\alpha_t \mathbf{x}_t} + \textcolor{green}{\beta_t\mathbf{\epsilon}_{\mathbf{\theta}}\left( \mathbf{x}_t, t \right)} + c_t\mathbf{\epsilon}$, partial backpropagation only utilizes the gradients of the red part ($\alpha_t \mathbf{x}_t$), truncating the green part (${\beta_t\mathbf{\epsilon}_{\mathbf{\theta}}\left( \mathbf{x}_t, t \right)}$). This inevitably introduces gradient errors, leading to unstable optimization. (d) Our method, using the few-step diffusion model to construct denoising shortcuts, facilitates complete gradient backpropagation through the entire denoising chain.}
\label{fig:fine-tune-baseline}
\end{figure*}
\section{Introduction}
\label{sec:introduction}

Diffusion models \cite{ho2020denoising, dhariwal2021diffusion, song2020score, song2020denoising, nichol2021improved, ho2022classifier, karras2022elucidating, rombach2022high, balaji2022ediffi} have established themselves as a pioneering approach in generative modeling, demonstrating exceptional prowess in applications such as photo-realistic text-to-image synthesis. However, the maximum likelihood training objective of these models, which aims to model the training data distribution accurately, can often conflict with downstream goals like aesthetics, fairness, safety, and text-to-image alignment. Therefore, aligning text-to-image diffusion models with human preferences has emerged as a pivotal and practical task.

Directly supervised fine-tuning on small-scale, human-curated datasets, \emph{e.g.}, LAION Aesthetics \cite{schuhmann2022laoin}, presents a straightforward solution. However, the prohibitive cost of data collection and the rapid obsolescence of datasets, particularly in terms of resolution compatibility with the latest text-to-image models, make this approach impractical.

Emulating the successful application of Reinforcement Learning from Human Feedback (RLHF) \cite{griffith2013policy, christiano2017deep, ziegler2019fine, stiennon2020learning, ouyang2022training} in Large Language Models (LLMs), several studies \cite{lee2023aligning, wu2023human, dong2023raft, black2023training, fan2024reinforcement} have experimented with Reinforcement Learning (RL) techniques to align diffusion models with a reward function. Despite promising performance enhancements in specific domains, RL-based methods are notorious for their high-variance gradients, leading to inefficiencies and limited adaptability to diverse prompts.

Recently, backpropagation-based methods \cite{clark2024directly, prabhudesai2023aligning, wu2024deep, xu2024imagereward} have sought to align diffusion models with reward functions using end-to-end backpropagation of the reward gradient through the denoising chain, showing potential. However, these strategies face challenges arising from the lengthy denoising chain, which demands considerable computational resources and is susceptible to gradient explosion. As illustrated in Fig.~\ref{fig:fine-tune-baseline} (b), \cite{clark2024directly, xu2024imagereward} have made strides by truncating backpropagation to focus on the latter part of the denoising chain. However, they overlook direct supervision in the early stages, leading to suboptimal alignment with text prompts. As illustrated in Fig.~\ref{fig:fine-tune-baseline} (c), \cite{prabhudesai2023aligning, wu2024deep} truncate a portion of backpropagation within the denoising chain by deactivating some gradients and introduce gradient checkpointing, enabling gradient backpropagation to the early stage of the denoising chain. However, they are time-consuming and introduce gradient bias, leading to optimization instability.

This paper revisits the issue of excessively long denoising chain and propose an alternative approach, employing the shorter denoising chain to facilitate full gradient backpropagation throughout the entire denoising chain. 

In this paper, we introduce Shortcut-based Fine-Tuning (\textsc{ShortFT}), which leverages the few-step diffusion model to construct a denoising \texttt{shortcut} to \textbf{fine-tune the foundational model} (\emph{e.g.}, SD 1.5), inspired by recent trajectory-preserving diffusion distillation methods \cite{salimans2022progressive, song2023consistency, kim2023consistency, ren2024hyper}. This technique enables us to bypass the original lengthy denoising chain and complete the inference, creating a shortcut-based denoising chain of shorter length. In addition, we construct a timestep-aware LoRA as an expert LoRA ensemble, based on the intriguing temporal dynamics exhibited by the text-to-image diffusion model during the denoising process. This approach increases the number of trainable parameters without increasing cost in the inference phase, enabling faster convergence and improved performance. We also devise a custom progressive training strategy to mitigate training inference bias introduced by using denoising shortcuts during the training stage.

Extensive quantitative and qualitative analyses demonstrate that \textsc{ShortFT} can be effectively applied to various reward functions and architectures, significantly enhancing alignment performance. Furthermore, \textsc{ShortFT}, benefiting from the short denoising chain and without the need for {gradient checkpointing}, is particularly efficient, learning faster than DRTune \cite{wu2024deep}, the current most efficient method.

\begin{figure*}
\centering
\setlength{\belowcaptionskip}{-0.3cm}
\setlength{\abovecaptionskip}{0.2cm}
\includegraphics[width=1.\linewidth]{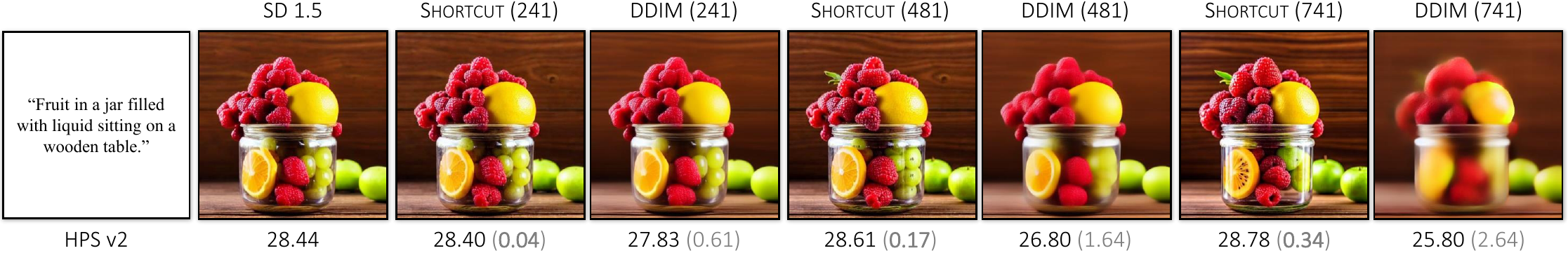}
\caption{\textbf{Denoising shortcut.} The trajectory-preserving few-step diffusion model naturally introduces denoising \texttt{shortcut}, allowing for flexible skipping within the denoising chain while still ensuring high-quality and consistent image synthesis. The 4-step Hyper-SD distilled from SD 1.5 is used in our experiments. ${\textsc{Shortcut}}(i)$ denotes completing the denoising process from timestep $i$ to $0$ using the few-step diffusion model. In addition, we also provide the well-known one-step denoising results $\text{DDIM}(i)$, where $\text{DDIM}(i)$ denotes performing a one-step denoising operation from timestep $i$ to $0$. Intuitively, the output of $\text{DDIM}(i)$ is more blurred, lacking accurate structure and texture details, while the output of ${\textsc{Shortcut}}(i)$ is closer to the original output of SD 1.5. Furthermore, we provide their corresponding HPS v2 scores, with the absolute value of the deviation from the score corresponding to SD 1.5 provided in parentheses, ${\textsc{Shortcut}}(i)$ exhibits smaller deviations. These observations collectively indicate the reliability and validity of the denoising shortcut.}
\label{fig:denoising_shortcut}
\end{figure*}

\section{Related Work}
\label{sec:related_work}

\subsection{Alignment of diffusion models}

\noindent Diffusion models \cite{ho2020denoising, dhariwal2021diffusion, song2020score, song2020denoising} have become a dominant force in generative modeling, showing exceptional performance in diverse applications \cite{nichol2021improved, ho2022classifier, karras2022elucidating, rombach2022high, balaji2022ediffi, xue2023raphael, gu2023matryoshka, guo2024initno, guo2024i4vgen, zhang2023adding}. However, certain misalignments with human intentions can arise. Recent research, fueled by the successful alignment of large language models, has sought to align diffusion models with human expectations and preferences.

\noindent \textbf{Fine-tuning via data augmentation.} Several studies \cite{huang2023t2i, wu2023human, lee2023aligning, dong2023raft, dai2023emu} have explored altering the training data distribution for fine-tuning diffusion models on visually compelling and textually coherent data, which has led to improved results. Other methods \cite{betker2023improving, segalis2023picture} involve re-captioning pre-collected web images to enhance textual precision.

\noindent \textbf{Fine-tuning via reward models.} Reward models \cite{kirstain2023pick, lee2023aligning, wu2023human, wu2023human2, xu2024imagereward, schuhmann2022laoin} are employed to emulate human preferences given an input prompt and generated images. Several approaches have attempted to integrate these signals to augment text-to-image generation. A significant direction is the utilization of reinforcement learning-based algorithms \cite{black2023training, deng2024prdp, chen2024enhancing, fan2023reinforcement, zhang2024large} for fine-tuning text-to-image diffusion models in alignment with these rewards, \cite{rafailov2024direct, wallace2024diffusion, yang2024using, liang2024step, yang2024dense, li2024aligning} bypass it entirely with Direct Preference Optimization. However, these methods are costly and have high gradient variance, leading to inefficiency and limited adaptability to diverse prompts. Consequently, backpropagation-based techniques \cite{clark2024directly, prabhudesai2023aligning, xu2024imagereward, wu2024deep} have been explored, which directly fine-tune diffusion models using differentiable rewards \cite{kirstain2023pick, wu2023human, wu2023human2, xu2024imagereward, schuhmann2022laoin}.

The challenge of backpropagation-based strategies stem from the lengthy denoising chain, which often requires numerous denoising operations (\emph{e.g.}, 50 for DDIM), corresponding to a long backpropagation chain. This process incurs substantial time and memory costs and is prone to gradient explosion. To mitigate this issue, \cite{clark2024directly, xu2024imagereward} truncate backpropagation by concentrating on the latter part of the denoising chain. While these approaches yields some improvements, they neglect direct supervision in the early stage of the denoising chain, leading to less precise alignment with text prompts. \cite{prabhudesai2023aligning, wu2024deep} truncate part of the backpropagation within the denoising chain by deactivating some gradients. By employing gradient checkpointing, they allows for the propagation of gradients to the early stages of the denoising chain. Despite their merits, these techniques can be computationally demanding and induce gradient bias, resulting in optimization instability.

Different from previous methods, this paper revisits the fundamental challenge of excessively long denoising chain and proposes an alternative approach: leveraging the shorter denoising chain to facilitate full gradient backpropagation throughout the denoising chain.

\begin{figure*}
\centering
\setlength{\belowcaptionskip}{-0.4cm}
\setlength{\abovecaptionskip}{0.2cm}
\includegraphics[width=0.99\linewidth]{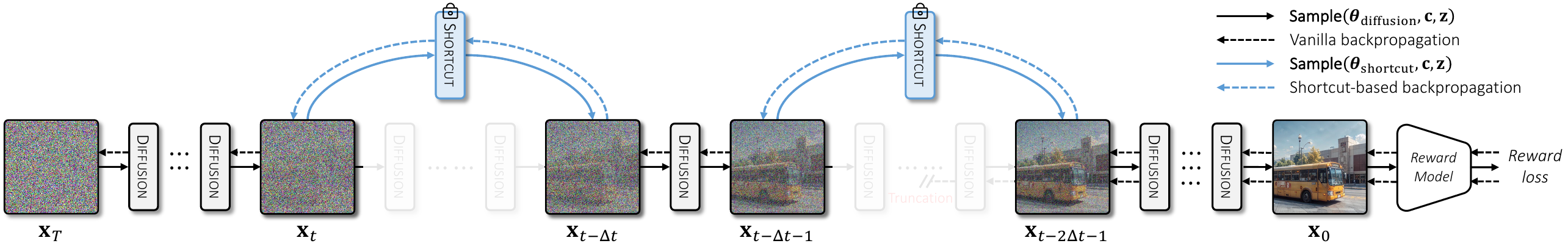}
\caption{\textbf{Illustration of \textsc{ShortFT}.} The core of the proposed method is the Shortcut-based Fine-Tuning (\textsc{ShortFT}), which leverages the trajectory-preserving few-step diffusion model as the \texttt{shortcut} (identified as blue arrow) to achieve direct end-to-end backpropagation through the diffusion sampling process, fine-tuning the parameters of the pre-trained diffusion model to align it with the reward function.}
\label{fig:shorft_illustration}
\end{figure*}

\subsection{Diffusion distillation}

Existing methods for the distillation of diffusion models can be primarily classified into two categories: trajectory-preserving distillation \cite{salimans2022progressive, song2023consistency, kim2023consistency, ren2024hyper} and trajectory-reformulating distillation \cite{sauer2024adversarial,kang2024distilling,yin2024one,lin2024sdxl,sauer2024fast}. The former aims to preserve the original denoising trajectory dictated by an ordinary differential equation (ODE), while the latter focuses on leveraging the denoising endpoint as the main supervision, disregarding the intermediate trajectory steps. This paper focuses on trajectory-preserving distillation, supporting the establishment of a denoising \texttt{shortcut} within the complete denoising chain.

\subsection{Fine-tuning few-step diffusion models}
\label{sec:fine-tuning-few-step-diffusion-models}

Existing works \cite{li2024reward, li2024t2v, li2024t2v2} have successfully explored fine-tuning few-step diffusion models. While they share similarities with our research in utilizing few-step diffusion models, a crucial difference is that we focus on fine-tuning the foundational models. In contrast to the foundational models, few-step diffusion models suffer from performance degradation and reduced capacity caused by the distillation process, leading fine-tuning them suboptimal (see Sec.~\ref{sec:more_results}).

\section{\textsc{ShortFT}}
\label{sec:approach}

Our method, Shortcut-based Fine-Tuning (\textsc{ShortFT}), capitalizes on the trajectory-preserving few-step diffusion model as a \texttt{shortcut} to achieve direct end-to-end backpropagation through the diffusion sampling process. This approach fine-tunes the parameters of the pre-trained diffusion model to align it with the reward function.

\subsection{Problem formulation}

In line with \cite{clark2024directly, prabhudesai2023aligning, wu2024deep}, \textsc{ShortFT} focuses on fine-tuning the parameters $\bm{\theta}$ of pre-trained diffusion models to maximize the differentiable reward function $\mathcal{R}(\cdot)$ for generated images. This can be formally represented as in Eq. \ref{eq:problem-formulation}:
\begin{equation}
\begin{aligned}
\label{eq:problem-formulation}
    J\left( \bm{\theta} \right) = \mathbb{E}_{\mathbf{c}, \mathbf{x}_T \sim \mathcal{N}\left( \mathbf{0}, \mathbf{1} \right)}\left[ \mathcal{R}\left( \mathsf{Sample}\left( \bm{\theta}, \mathbf{c}, \mathbf{x}_T \right), \mathbf{c} \right) \right],
\end{aligned}
\end{equation}
where $\mathsf{Sample}\left( \bm{\theta}, \mathbf{c}, \mathbf{x}_T \right)$ represents the denoising process for the timestep $t = T \to 0$ with prompt condition $\mathbf{c}$.

Consistent with \cite{clark2024directly, prabhudesai2023aligning, wu2024deep}, Eq. \ref{eq:problem-formulation} is resolved by calculating $\nabla \mathcal{R}\left( \mathsf{Sample}\left( \bm{\theta}, \mathbf{c}, \mathbf{x}_T \right), \mathbf{c} \right)$ and employing gradient ascent. The computation of this gradient necessitates backpropagation through multiple diffusion models in the denoising chain, akin to backpropagation through time in recurrent neural networks.

\subsection{Denoising shortcut}
\label{sec:denoising-shortcut}

The recent emergence of a series of diffusion-aware distillation algorithms \cite{salimans2022progressive, song2023consistency, kim2023consistency, ren2024hyper} has been instrumental in mitigating the computational burden associated with the multi-step inference process of diffusion models. These algorithms can be roughly classified into two categories: trajectory-preserving distillation \cite{salimans2022progressive, song2023consistency, kim2023consistency, ren2024hyper} and trajectory-reformulating distillation \cite{sauer2024adversarial,kang2024distilling,yin2024one,lin2024sdxl,sauer2024fast}.

Among these, trajectory-preserving few-step diffusion models naturally introduce a denoising \texttt{shortcut}, which allows for flexible skipping within the denoising chain while still ensuring high-quality and consistent image synthesis. As demonstrated in Fig.~\ref{fig:denoising_shortcut}, the integration of these shortcuts into the denoising chain significantly reduces the total number of denoising steps, thereby shortening the length of the denoising chain. This key insight paves the way for efficient and effective end-to-end backpropagation.

\begin{figure}
\centering
\setlength{\belowcaptionskip}{-0.2cm}
\setlength{\abovecaptionskip}{0.2cm}
\includegraphics[width=0.95\linewidth]{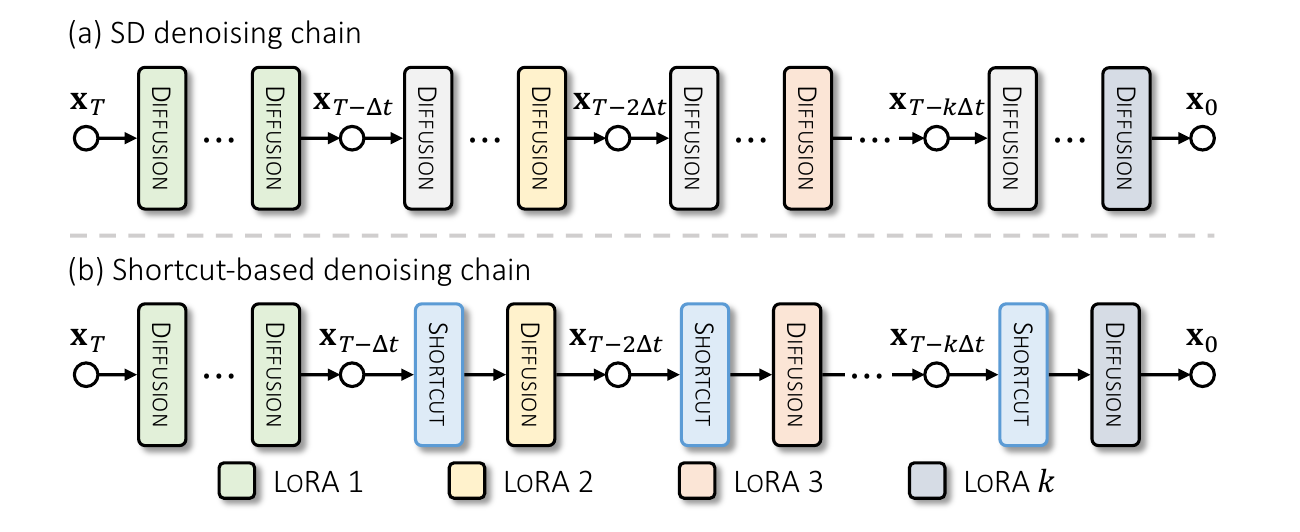}
\caption{\textbf{Illustration of timestep-aware LoRA.} (a) Vanilla SD denoising chain; (b) Shortcut-based denoising chain. In accordance with the interesting time dynamics in the text-to-image diffusion model denoising process revealed by \cite{balaji2022ediffi}, different from existing methods that share the same LoRA parameters at all timesteps, we introduce time-step aware LoRA, which effectively increases the capacity of the diffusion model and accelerates the convergence of training, without increasing the computational cost during the inference stage.}
\label{fig:timestep_aware_lora}
\end{figure}

\subsection{Shortcut-based fine-tuning}

\noindent \textbf{Insight.} The naive optimization of Eq.~\ref{eq:problem-formulation} through backpropagation necessitates the construction of the complete denoising chain: $\{\mathbf{x}_T, \cdots, \mathbf{x}_t, \cdots, \mathbf{x}_{t - \Delta t}, \mathbf{x}_{t - \Delta t - 1}, \cdots,$ $ \mathbf{x}_{t - 2\Delta t - 1}, \cdots, \mathbf{x}_0\}$. This process involves the storage of intermediate activations linked to each neural layer and each denoising timestep within GPU VRAM, which is not feasible due to memory constraints. Furthermore, the typical length of a denoising chain is approximately 50, which results in an overly long backpropagation chain that can lead to issues of gradient explosion. 

The key insight of \textsc{ShortFT} is rooted in the utilization of the trajectory-preserving few-step diffusion model to construct the denoising \texttt{shortcut}, significantly reducing the length of the denoising chain. As elaborated in Sec. \ref{sec:denoising-shortcut}, this denoising {shortcut} enables us to bypass a substantial number of denoising timesteps, leading to create a more streamlined and efficient denoising chain: $\{\mathbf{x}_T, \cdots,$ $\textcolor{iccvblue}{\mathbf{x}_t, \mathbf{x}_{t - \Delta t}}, \textcolor{iccvblue}{\mathbf{x}_{t - \Delta t - 1}, \mathbf{x}_{t - 2\Delta t - 1}}, \cdots, \mathbf{x}_0\}$. 

As depicted in Fig. \ref{fig:shorft_illustration}, such a design allows for the direct implementation of reward supervision at the early stages of the denoising chain, and facilitates full gradient backpropagation throughout the denoising chain.

\noindent \textbf{Shortcut-based denoising chain.} As illustrated in Fig.~\ref{fig:timestep_aware_lora}, a vanilla SD denoising chain comprises a sequence of step-by-step denoising operations that transform the input noise $\mathbf{z}_T$ into the output image $\mathbf{z}_0$. This process typically necessitates numerous denoising steps. By harnessing the denoising \texttt{shortcut}, we are able to construct a shortcut-based denoising chain and fine-tune the diffusion model through end-to-end backpropagation.

\begin{figure}
\centering
\setlength{\belowcaptionskip}{-0.2cm}
\setlength{\abovecaptionskip}{0.2cm}
\includegraphics[width=0.95\linewidth]{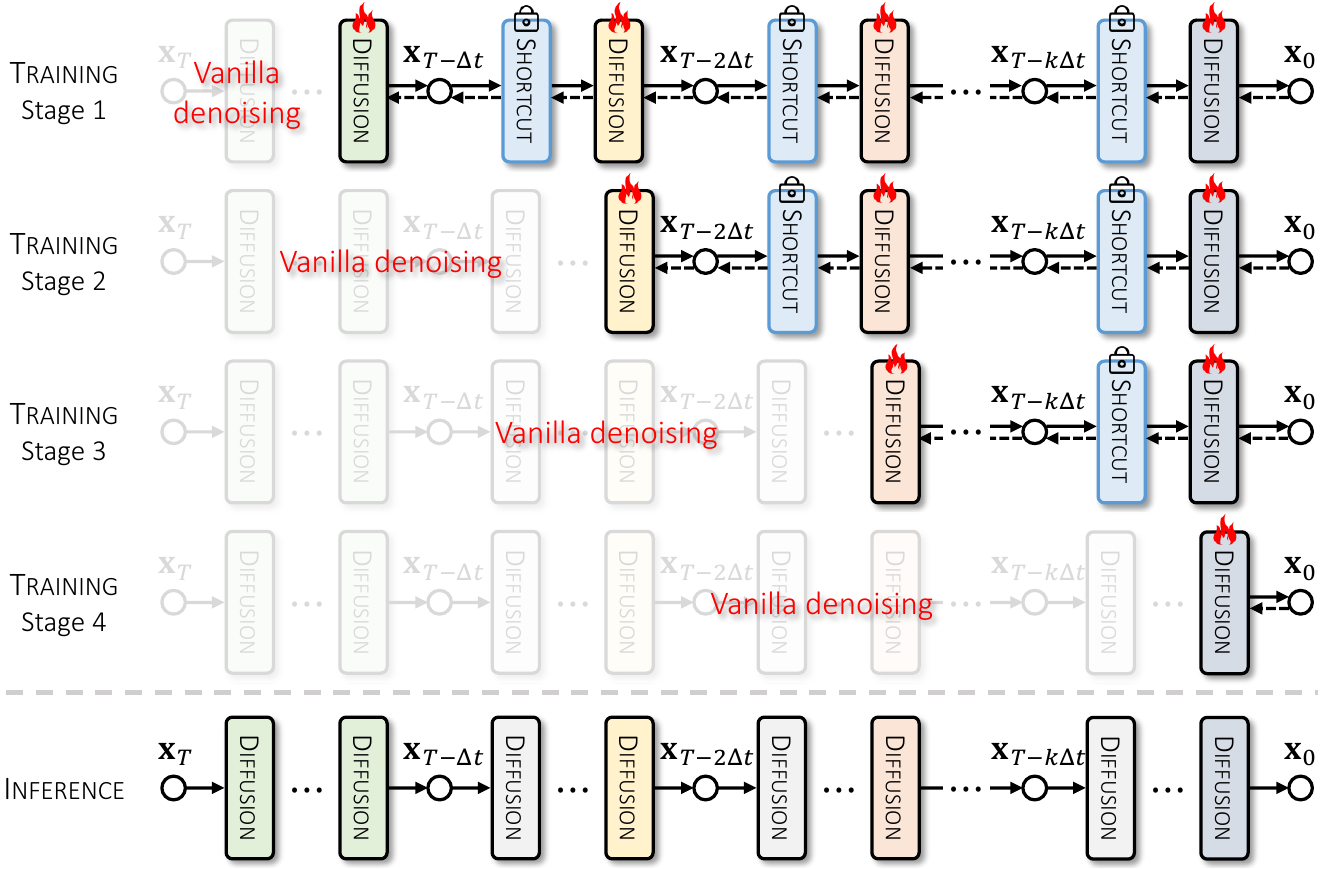}
\caption{\textbf{Illustration of progressive training strategy.} Corresponding to the time-step aware LoRA design, we develop the progressive training strategy, which eliminates the training-inference gap introduced by the shortcut-based fine-tuning.}
\label{fig:progressive_training}
\end{figure}

\noindent \textbf{Timestep-aware LoRA.} Instead of fine-tuning the full weights of the original diffusion model, Low-Rank Adaptation (LoRA) \cite{hu2021lora} preserves the weights of the pre-trained model and introduces new low-rank weight matrices alongside the original model weights. The contributions of these matrices are summed to generate the final outputs. Specifically, each linear layer of the UNet of SD is modified from $\mathbf{h} = \mathbf{W} \mathbf{x}$ to $\mathbf{h} = \mathbf{W} \mathbf{x} + \mathbf{B}\mathbf{A}\mathbf{x}$, where $\mathbf{W}\in \mathbb{R}^{d\times d}$, $\mathbf{B}\in \mathbb{R}^{d\times k}$, $\mathbf{A}\in \mathbb{R}^{k\times d}$, and $k \ll d$. LoRA considerably reduces the number of parameters to be optimized, thereby decreasing the memory requirements for fine-tuning.

Moreover, \cite{balaji2022ediffi} uncovers intriguing temporal dynamics during the denoising process of the text-to-image diffusion model. In the initial sampling stage, the model largely depends on the text prompts to guide the sampling process. As the generation progresses, the model gradually leans on visual features to denoise the image. This indicates that sharing LoRA parameters (standard practice in the existing methods) throughout the entire denoising process may not be optimal and may fail to capture the distinct patterns that emerge during denoising. Therefore, in contrast to \cite{clark2024directly, prabhudesai2023aligning, wu2024deep, xu2024imagereward} which share the same LoRA parameters for all timesteps, we introduce timestep-aware LoRA. This design effectively increases the capacity of the diffusion model and accelerates the convergence of training, without increasing computational cost in the inference phase.

Specifically, as depicted in Fig.~\ref{fig:timestep_aware_lora}, we initially divide the entire denoising chain into $k$ segments, $\Delta t=\lfloor \frac{T}{k} \rfloor$. Except for the first segment, we assign a corresponding LoRA to the last timestep of each subsequent segment. For the first segment, we adopt the methodology of \cite{clark2024directly} and share the same LoRA parameters across all timesteps. Notably, a sequence of continuous timesteps lacking LoRA exists in later segments, supporting the denoising shortcuts.

\noindent \textbf{Progressive training strategy.} Although the few-step diffusion model is capable of creating the denoising shortcut, bypassing the entire denoising chain, it inherently introduces errors in the output, meaning it still cannot be fully consistent with the output of SD, particularly in fine details, as shown in Fig.~\ref{fig:denoising_shortcut}. This incongruity can lead to a training-inference gap, resulting in suboptimal results. To mitigate this, we design a progressive training strategy.

As illustrated in Fig.~\ref{fig:progressive_training}, in accordance with the design of timestep-aware LoRA, we divide the \textsc{ShortFT} training process into $k$ stages. For the $i$-th training stage, we optimize the weights of LoRA $i$ to LoRA $k$. For the $i$-th segment and preceding denoising processes, we retain the original denoising chain, while for the denoising processes post the $i$-th segment, we introduce the denoising shortcut, thereby shortening the depth of the backpropagation chain. Moreover, in line with \cite{clark2024directly}, we also employ the truncated backpropagation technique.

During inference, as displayed in Fig.~\ref{fig:progressive_training}, the denoising shortcut is bypassed, and the original denoising chain is used to generate the final output image.

\begin{figure*}
\centering
\setlength{\belowcaptionskip}{-0.4cm}
\setlength{\abovecaptionskip}{0.2cm}
\includegraphics[width=0.93\linewidth]{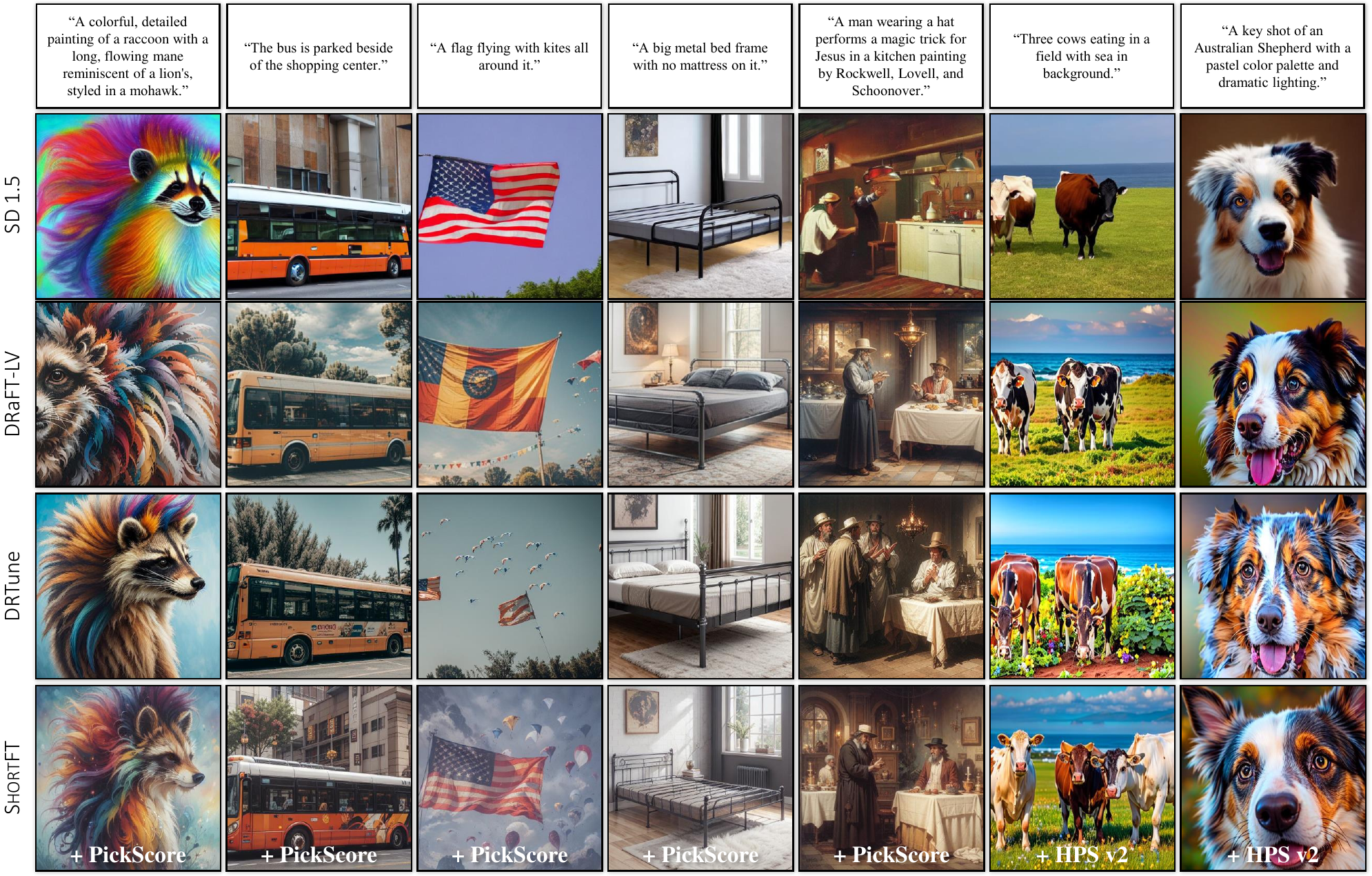}
\caption{\textbf{Qualitative comparison on PickScore and HPS v2.} Each image is generated with the same text prompt and random seed for all methods. Our method outperforms existing methods in both text-image alignment and image quality.}
\label{fig:qualitative_comparison_pickscore_hps}
\end{figure*}

\begin{figure*}
\centering
\setlength{\belowcaptionskip}{-0.3cm}
\setlength{\abovecaptionskip}{0.2cm}
\includegraphics[width=0.93\linewidth]{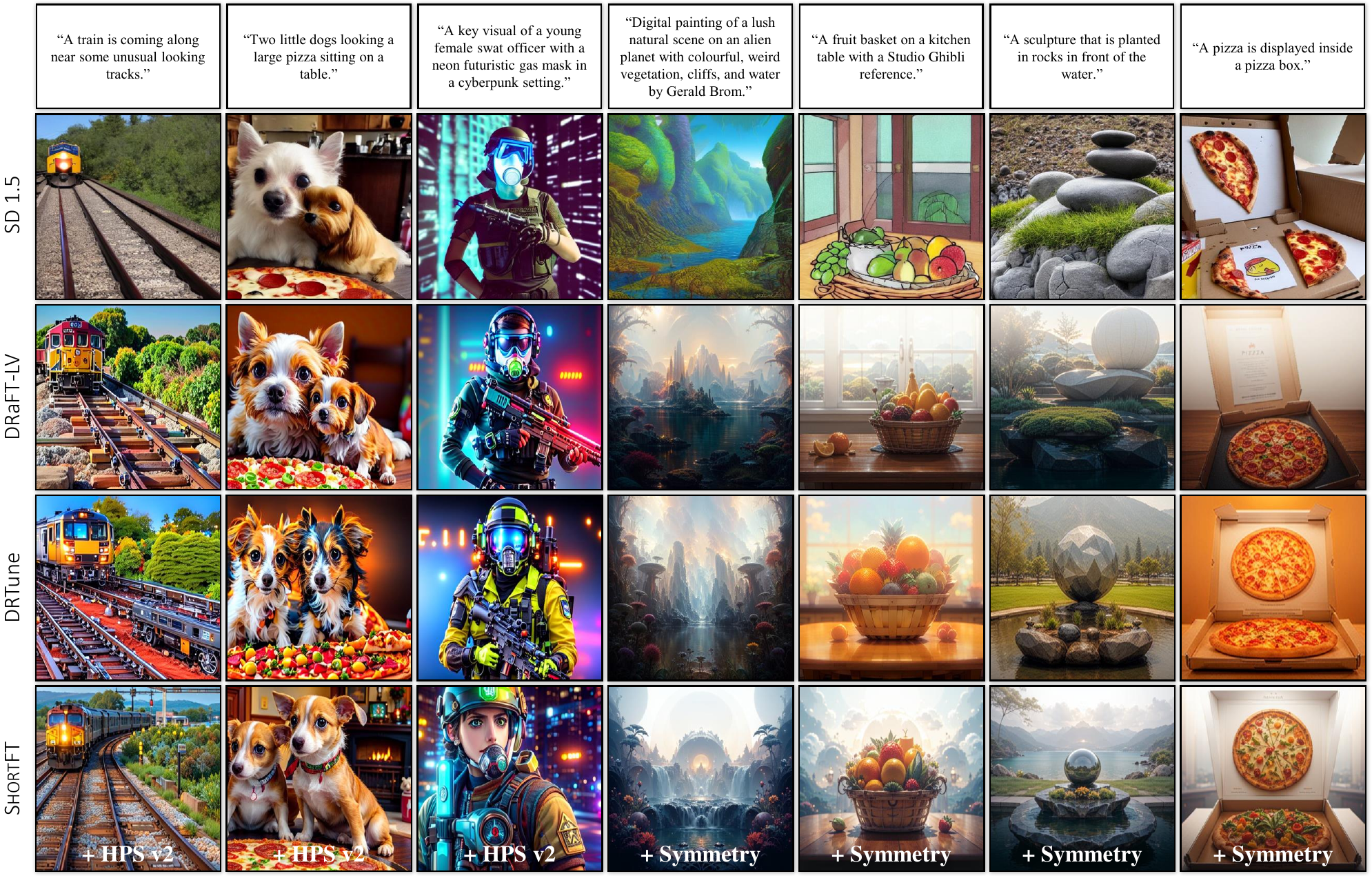}
\caption{\textbf{Qualitative comparison on HPS v2 and Symmetry.} Each image is generated with the same text prompt and random seed for all methods. Our method outperforms existing methods in both text-image alignment and image quality.}
\label{fig:qualitative_comparison_hps_symmetry}
\end{figure*}

\section{Experiments}
\label{sec:experiments}

\subsection{Experimental settings}

In our experiments, Stable Diffusion 1.5 serves as the foundational diffusion model. The DDIM schedule  \cite{song2020denoising} is employed to execute 50 steps of denoising, with a classifier-free guidance scale of 7.5.

\noindent \textbf{Shortcut.} \textsc{ShortFT} shortens the denoising chain by bypassing certain steps within the chain. Hence, the selection of few-step diffusion models is focused predominantly on methods that employ trajectory-preserving distillation algorithms. To accommodate the proposed time-aware LoRA, one-step diffusion models are avoided, whose output image quality is also relatively inferior. Specifically, 4-step Hyper-SD \cite{ren2024hyper}, distilled from SD 1.5, is utilized to construct the denoising shortcut. The value of $k$ is set to 4, and the timesteps configured for LoRA are \{761, 501, 261, 1\}. Consequently, the denoising shortcuts are executed separately between timesteps 741 to 501, timesteps 481 to 261, and timesteps 241 to 1.

\noindent \textbf{Timestep-aware LoRA.} As suggested by \cite{clark2024directly}, LoRA is applied to both the feedforward and attention layers in the UNet. The LoRA rank is set to 128. Furthermore, we adopt a stepwise stacking approach, where for LoRA $i$, we introduce a new LoRA branch on top of LoRA ${i-1}$.

\noindent \textbf{Reward functions.} The proposed method is evaluated using three reward functions: Human Preference Score v2 (HPS v2) \cite{wu2023human2}, PickScore \cite{kirstain2023pick}, and Symmetry \cite{wu2024deep}. HPS v2 and PickScore capture human preference for images based on input prompts, while Symmetry encourages images to have horizontal symmetry features. Different from \cite{wu2024deep}, which utilizes CLIPScore \cite{radford2021learning} as a regularization term, our experiment amalgamates HPS v2 and PickScore in a ratio of 1:10 to function as a joint regularization term. This approach has demonstrated superior performance in terms of text-image alignment and overall image quality.

\noindent \textbf{Experimental details.} All experiments are conducted using 2 A800 GPUs and the AdamW optimizer with $\beta_1 = 0.9$, $\beta_2 = 0.999$, and a weight decay of $0.1$. \textsc{ShortFT} is performed with a batch size of 128 and a constant learning rate of $5\times 10^{-5}$. During training, the pre-trained SD parameters are converted to bfloat16 to reduce memory usage, while the LoRA parameters under training remain in float32. Gradient checkpointing is not required.

\begin{table}
\small
\centering
\setlength{\belowcaptionskip}{-0.3cm}
\setlength{\abovecaptionskip}{0.2cm}
\scalebox{0.95}{
\begin{tabular}{l|ccc}
\toprule
\multicolumn{1}{c|}{\bf Method} & {\bf HPS v2}$^\uparrow$ & {\bf PickScore}$^\uparrow$ & {\bf Symmetry}$^\downarrow$ \\
\midrule
\midrule
SD 1.5 \cite{rombach2022high} & 26.91 & 20.46 & 0.853 \\
DRaFT-LV \cite{clark2024directly} & 33.13 & 23.35 & {0.418} \\
DRTune \cite{wu2024deep} & 32.79 & 23.22 & 0.207 \\
\midrule
\textsc{ShortFT} & {\bf 33.88} & {\bf 24.16} & {\bf 0.138} \\
\bottomrule
\end{tabular}}
\caption{\textbf{Objective evaluation.} Our method performs over other counterparts, under the same computational cost.}
\label{tab:objective-evaluation}
\end{table}

\noindent \textbf{Datasets.} We compare \textsc{ShortFT}-finetuned diffusion models to those of the state-of-the-art counterparts on the Human Preference Score v2 dataset (HPDv2) \cite{wu2023human2}. The final reward is computed on the 400 prompts from the test split. Following \cite{wu2024deep}, for fair comparison, we evaluate all methods using the same computational budget. Specifically, all methods are trained for six hours on 2 A800 GPUs.

\subsection{Qualitative comparison}
\label{sec:qualitative_comparison}

Fig.~\ref{fig:qualitative_comparison_pickscore_hps} and \ref{fig:qualitative_comparison_hps_symmetry} present the quantitative comparison of our results against those of representative methods, including current state-of-the-art techniques, under same text prompts and random seeds. SD 1.5 suffers from low-quality image generation. DRaFT \cite{clark2024directly}, while proficient at managing local image details, struggles to effectively handle global layouts and the optimization of the symmetry reward function. Similarly, DRTune \cite{wu2024deep} is constrained by gradient inaccuracies, which contribute to instability during training and a deficiency in managing complex semantics. One particular area of weakness is the synthesis of images with specific counts of objects, such as accurately depicting a given number of cows. In contrast, \textsc{ShortFT} exhibits enhanced capabilities in generating images that are both visually realistic and semantically faithful, outperforming over other counterparts.

\subsection{Quantitative comparison}
\label{sec:quantitative_comparison}

\noindent \textbf{Objective evaluation.} Table~\ref{tab:objective-evaluation} shows the quantitative results achieved on the Human Preference Score v2 benchmark \cite{wu2023human2}, where the proposed method outperforms the other approaches, clearly demonstrating its effectiveness. It is important to highlight that both our method and DRTune \cite{wu2024deep} strategically employ backpropagation of the reward gradient to the initial stages of the denoising chain. This intentional design choice significantly enhances the Symmetry score performance when compared to alternative methods. Despite this, DRTune continue to grapple with the challenge of gradient bias. In contrast, our method significantly mitigates this issue, delivering superior performance.

\begin{figure}
\centering
\setlength{\belowcaptionskip}{-0.6cm}
\setlength{\abovecaptionskip}{0.2cm}
\includegraphics[width=0.9\linewidth]{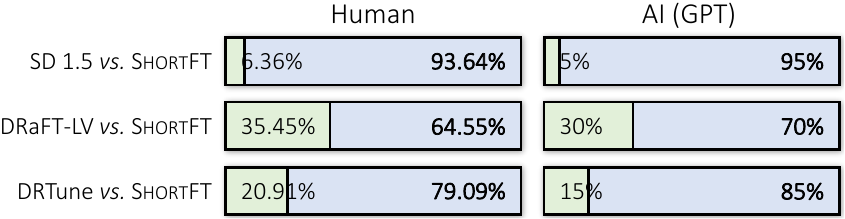}
\caption{\textbf{Human and AI preference evaluation} against current methods. \textsc{ShortFT} performs over other counterparts.}
\label{fig:user_study}
\end{figure}

\noindent \textbf{User study.} A subjective user study comprising 11 volunteers is conducted, with five possessing expertise in image processing and the remaining participants having no background in computer vision or graph. Participants are tasked to select the most visually attractive and semantically accurate image among those generated by our method and current state-of-the-art techniques. Each participant has 10 questions for each pair of comparisons. Furthermore, an MLLM-assisted evaluation is employed using {GPT-4V}. We make 20 queries to GPT-4V for each pair of comparisons. More details are provided in the Appendix. As depicted in Fig.~\ref{fig:user_study}, the results exhibit a significant inclination towards \textsc{ShortFT} in comparison to other techniques.

\subsection{More results}
\label{sec:more_results}

\noindent \textbf{\textsc{ShortFT}, 10k training step.} Due to the shorter denoising and backpropagation chains, our method achieves superior performance under the same computational budget. Furthermore, to validate the upper bound of our approach, following the protocol in \cite{clark2024directly}, we conduct the full training process using HPS v2 reward function on HPDv2, comprising 10k training steps, and evaluate it on the corresponding benchmark. The obtained HPS v2 score of \emph{35.97} surpasses the reported score for DRaFT-LV in \cite{clark2024directly}.

\begin{table}[ht]
\small
\centering
\setlength{\belowcaptionskip}{-0.2cm}
\setlength{\abovecaptionskip}{0.2cm}
\scalebox{0.9}{
\begin{tabular}{l|cc}
\toprule
\multicolumn{1}{c|}{\bf Method} & {\bf Tuning Hyper-SD} & {\bf Tuning SD 1.5} \\
\midrule
\midrule
HPS v2$^\uparrow$ & 32.92 & \textbf{35.97} \\
\bottomrule
\end{tabular}}
\caption{\textbf{Objective evaluation} on tuning SD 1.5 and Hyper-SD.}
\label{tab:clarification}
\end{table}

\begin{figure}[ht]
\centering
\setlength{\belowcaptionskip}{-0.5cm}
\setlength{\abovecaptionskip}{0.2cm}
\includegraphics[width=1.\linewidth]{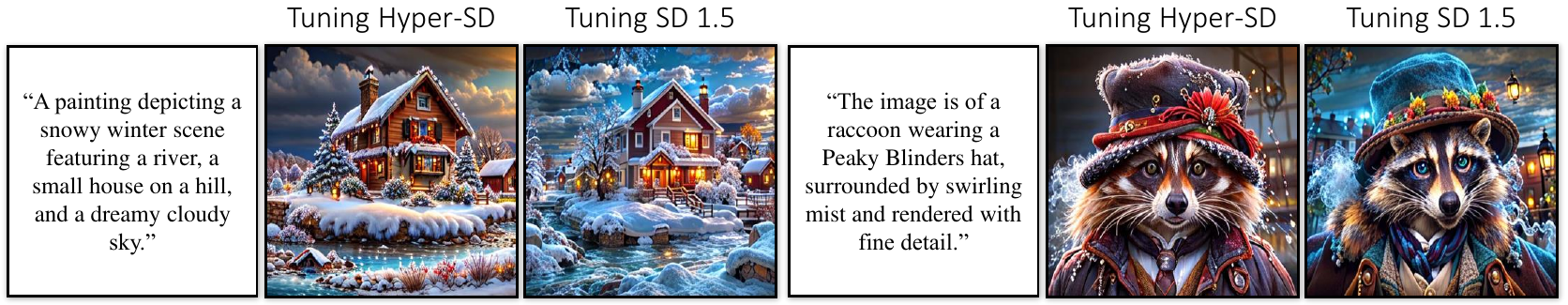}
\caption{\textbf{Qualitative comparison} on tuning SD 1.5 (\textsc{ShortFT}) and Hyper-SD. Fine-tuning SD 1.5 significantly outperforms fine-tuning Hyper-SD, with the former enjoying more exquisite details.}
\label{fig:clarification}
\end{figure}

\noindent \textbf{Fine-tuning SD vs. Hyper-SD.} As mentioned in Sec.~\ref{sec:fine-tuning-few-step-diffusion-models}, \cite{li2024reward, li2024t2v, li2024t2v2} explore fine-tuning few-step diffusion models and have achieved certain successes. However, compared to the foundational model, few-step diffusion models face performance degradation caused by the distillation process. Fine-tuning the few-step diffusion model is actually suboptimal compared to fine-tuning the foundational model. Furthermore, following two strategies, we separately conduct the training processes on the HPDv2 using the HPS v2 reward function. As shown in Table~\ref{tab:clarification} and Fig.~\ref{fig:clarification}, fine-tuning SD 1.5 (\textsc{ShortFT}) significantly outperforms fine-tuning Hyper-SD, with the former enjoying more exquisite details.

\begin{figure}[ht]
\centering
\setlength{\belowcaptionskip}{-0.2cm}
\setlength{\abovecaptionskip}{0.2cm}
\includegraphics[width=1.\linewidth]{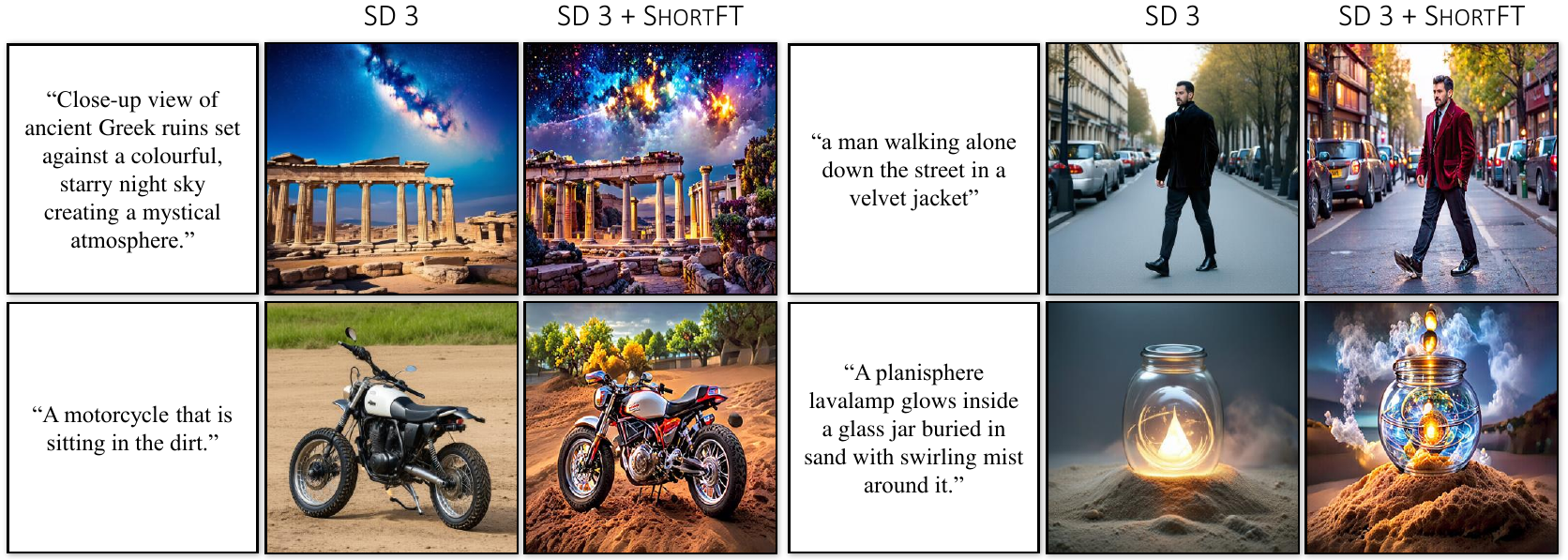}
\caption{\textbf{Example results} synthesized by \textsc{ShortFT} on SD 3.}
\label{fig:sd3}
\end{figure}

\noindent \textbf{Fine-tuning SD 3.} \textsc{ShortFT} is an architecture-agnostic fine-tuning strategy, applicable to both UNet-based (SD 1.5) and Transformer-based (SD 3) architectures. As illustrated in Fig.~\ref{fig:sd3}, \textsc{ShortFT} is also capable of mastering SD 3, where SD 3 aligns with HPS v2.

\begin{figure}[ht]
\centering
\setlength{\belowcaptionskip}{-0.2cm}
\setlength{\abovecaptionskip}{0.2cm}
\includegraphics[width=0.98\linewidth]{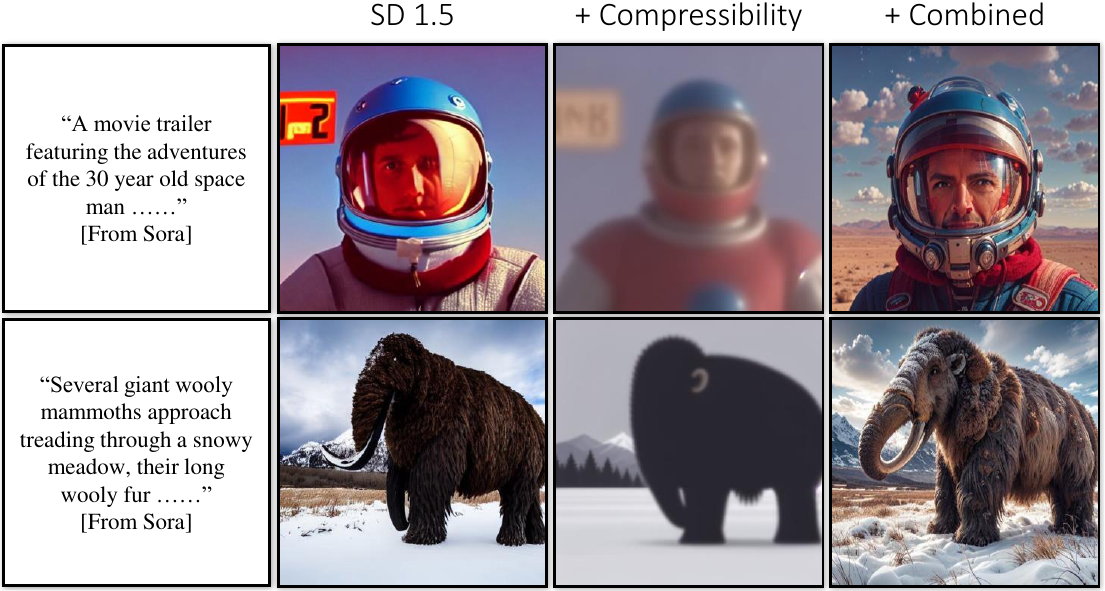}
\caption{\textbf{Generalization to wild text prompts} from Sora. Our method is capable of effectively handling the wild text prompts.}
\label{fig:generalization}
\end{figure}

\noindent \textbf{Other reward functions.} \textsc{ShortFT} exhibits remarkable versatility, demonstrating efficacy across a spectrum of reward functions, significantly improving the alignment performance and the quality and fidelity of the output. As shown in Fig.~\ref{fig:teaser} and~\ref{fig:generalization}, \textsc{ShortFT} not only accommodates HPS v2, PickScore, and Symmetry, but also exhibits proficiency in managing Compressibility and Combined reward.

\noindent \textbf{Generalization to wild text prompts.} As shown in Fig.~\ref{fig:generalization}, we present the qualitative results of prompt generalization. We found that using \textsc{ShortFT} to fine-tune the model on HPDv2 still enables effective handling of wild text prompts, enhancing the overall quality of the generated images.

\begin{table}
\small
\centering
\setlength{\belowcaptionskip}{-0.2cm}
\setlength{\abovecaptionskip}{0.2cm}
\scalebox{0.9}{
\begin{tabular}{l|ccc}
\toprule
\multicolumn{1}{c|}{\bf Method} & {\bf HPS v2}$^\uparrow$ & {\bf PickScore}$^\uparrow$ & {\bf Symmetry}$^\downarrow$ \\
\midrule
\midrule
w/o T-LoRA & 33.46 & 23.82 & 0.187 \\
w/o P-Training & 33.27 & 23.97 & 0.146 \\
\midrule
\textsc{ShortFT} & {\bf 33.88} & {\bf 24.16} & {\bf 0.138} \\
\bottomrule
\end{tabular}}
\caption{\textbf{Ablation study} on timestep-aware LoRA.}
\label{tab:ablation-study}
\end{table}

\begin{figure}
\centering
\setlength{\belowcaptionskip}{-0.4cm}
\setlength{\abovecaptionskip}{0.2cm}
\includegraphics[width=0.8\linewidth]{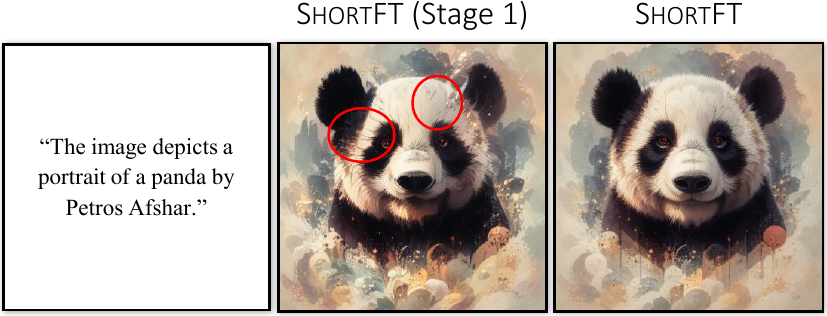}
\caption{\textbf{Ablation study} on progressive training strategy. The red circle marks the incoherent local details, \emph{i.e.}, unsmooth hair.}
\label{fig:ablation_progressive_training}
\end{figure}

\subsection{Ablation study}
\label{sec:ablation_study}

\noindent \textbf{On timestep-aware LoRA.} As shown in Table~\ref{tab:ablation-study},  timestep-aware LoRA effectively increases the capacity of the diffusion model and accelerates the convergence of training. Under the same computational cost, the time-aware LoRA achieves better performance.

\noindent \textbf{On progressive training strategy.} The progressive training strategy is committed to eliminating the training-inference gap. As shown in Table~\ref{tab:ablation-study} and Fig.~\ref{fig:ablation_progressive_training}, directly integrating the LoRA parameters obtained from training stage 1 into the pre-trained diffusion model results in incoherent local details, resulting in worse results, which can be effectively handled by progressive training strategy.

\section{Conclusion}
\label{sec:conclusion}

In this paper, we propose a novel Shortcut-based Fine-Tuning (\textsc{ShortFT}), an advanced technique for aligning diffusion models with reward functions through end-to-end backpropagation in the denoising chain. While existing methods struggle with computational costs and the risk of gradient explosion, \textsc{ShortFT} leverages shorter denoising chains, markedly improving fine-tuning efficiency and effectiveness. Rigorous evaluations demonstrate that our method can be effectively applied to various reward functions, significantly enhancing alignment performance and surpassing state-of-the-art alternative solutions.

\section*{Acknowledgment} This work is supported by the National Key Research and Development Plan (2024YFB3309302).

{
    \small
    \bibliographystyle{ieeenat_fullname}
    \bibliography{main}
}

\end{document}